\documentclass{article}

\usepackage{amsmath,amsfonts,bm}

\def\eqref#1{equation~\ref{#1}}

\def\1{\bm{1}}

\def\mA{{\bm{A}}}

\def\mC{{\bm{C}}}
\def\mD{{\bm{D}}}

\def\mI{{\bm{I}}}

\def\mL{{\bm{L}}}
\def\mM{{\bm{M}}}

\def\mO{{\bm{O}}}

\def\mU{{\bm{U}}}

\def\mW{{\bm{W}}}
\def\mX{{\bm{X}}}
\def\mY{{\bm{Y}}}
\def\mZ{{\bm{Z}}}

\DeclareMathAlphabet{\mathsfit}{\encodingdefault}{\sfdefault}{m}{sl}
\SetMathAlphabet{\mathsfit}{bold}{\encodingdefault}{\sfdefault}{bx}{n}
\newcommand{\tens}[1]{\bm{\mathsfit{#1}}}

\def\tX{{\tens{X}}}
\def\tY{{\tens{Y}}}

\usepackage{url}
\usepackage{microtype}
\usepackage{graphicx}
\usepackage{booktabs} 
\usepackage{caption}
\usepackage{subcaption}

\usepackage{hyperref}

\usepackage{icml2020}

\usepackage{lipsum}
\usepackage{epstopdf}
\usepackage{amsmath}
\usepackage{mathrsfs}
\usepackage{amsthm}
\usepackage{multirow}
\usepackage{bm}

\newtheorem{Theorem}{Theorem}
 
\newtheorem{Remark}{Remark} 
\newtheorem{Property}{Property}

\icmltitlerunning{Spectral Nonlocal Block}

\begin{document}

\twocolumn[
\icmltitle{A Spectral Nonlocal Block for Deep Neural Networks}



\icmlsetsymbol{equal}{*}
\begin{icmlauthorlist}
\icmlauthor{Lei Zhu*}{to}
\icmlauthor{Qi She*}{to}
\icmlauthor{Lidan Zhang}{to}
\icmlauthor{Ping Guo}{to}
\end{icmlauthorlist}

\icmlaffiliation{to}{Intel Lab China, Beijing, China}


\begin{center}
    Intel Research Lab China, Beijing, China \\ \texttt{lei1.zhu@intel.com}\\
\end{center}

\icmlkeywords{Machine Learning, ICML}

\vskip 0.3in
]




\begin{abstract}
The nonlocal-based blocks are designed for capturing long-range spatial-temporal dependencies in computer vision tasks. Although having shown excellent performances, they lack the mechanism to encode the rich, structured information among elements in an image. 
In this paper, to theoretically analyze the property of these nonlocal-based blocks, we provide a unified approach to interpreting them, where we view them as a graph filter generated on a fully-connected graph. When the graph filter is approximated by Chebyshev polynomials, a generalized formulation can be derived for explaining the existing nonlocal-based blocks ($\mathit{e.g.,}$ nonlocal block, nonlocal stage, double attention block). Furthermore, we propose an efficient and robust spectral nonlocal block, which can be flexibly inserted into deep neural networks to catch the long-range dependencies between spatial pixels or temporal frames. Experimental results demonstrate the clear-cut improvements and practical applicabilities of the spectral nonlocal block on image classification (Cifar-10/100, ImageNet), fine-grained image classification (CUB-200), action recognition (UCF-101), and person re-identification (ILID-SVID, Mars, Prid-2011) tasks.

\end{abstract}

\section{Introduction}
Capturing the long-range spatial-temporal dependencies between spatial pixels or temporal frames plays a crucial role in the computer vision tasks. Convolutional neural networks (CNNs) are inherently limited by their convolution operators which are devoted to capture local features and relations, $\mathit{e.g.,}$ a $7 \times 7$ region, and are inefficient in modeling long-range dependencies. Deep CNNs model these dependencies, which commonly refers to enlarge receptive fields, via stacking multiple convolution operators. However, two unfavorable issues are raised in practice. Firstly, repeating convolutional operations comes with higher computation and memory cost as well as the risk of over-fitting~\cite{cost}. Secondly, stacking more layers cannot always increase the effective receptive fields~\cite{widelimit}, which indicates the convolutional layers may still lack the mechanism to efficiently model these dependencies.

\begin{figure*}[!htbp]
    \centering
    \includegraphics[width=0.90\textwidth]{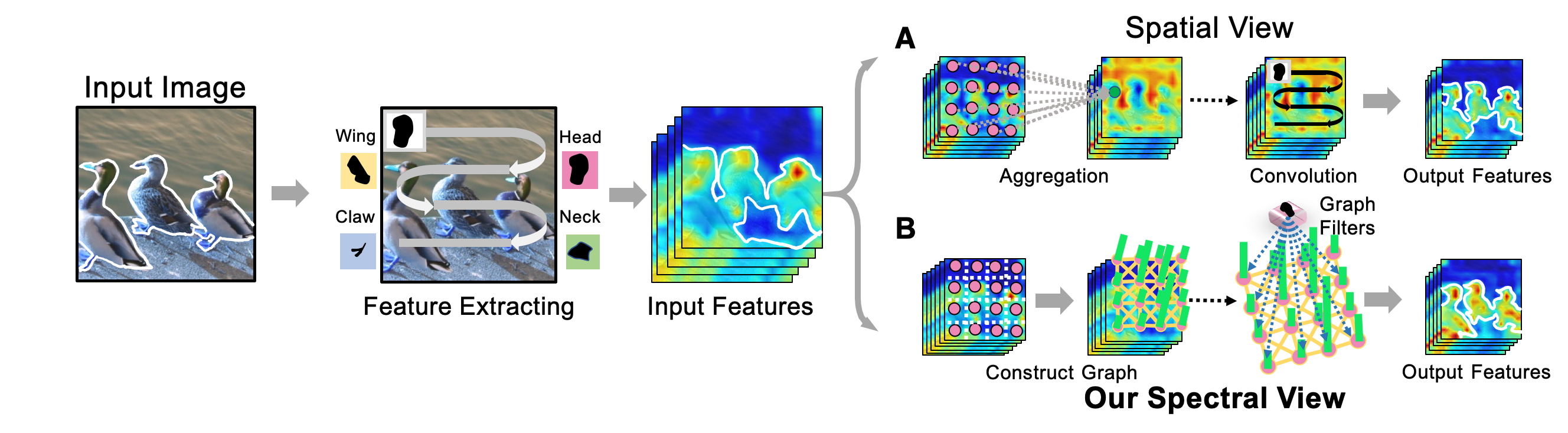}
    \vspace{-5mm}
    \caption{The spatial (\textbf{\sf{A}}) and spectral (\textbf{\sf{B}}) view of a nonlocal block. The pink dots indicate each patch in the feature map and the ``Aggregation" means calculating the weighted mean as the numerator of Eq.~(\ref{org_nlop}). The dotted arrows mean ``copy" and full arrows mean ``feed forward". The green bars are the node features and the length means their strength (best view in color).}
    \label{intro}
    \vspace{-2mm}
\end{figure*}

To address these issues, inspired by the classical ``nonlocal mean" method in image denoising field, \citet{nl} proposes the nonlocal block to encode the ``full-range" dependencies in one module by exploring the relation between pairwise positions. For each central position, the nonlocal block firstly computes the pairwise relations between the central position and all positions to form an attention map, and then aggregates the features of all positions by weighted mean according to the attention map. The aggregated features are filtered by the convolutional operator and finally added to the features of each central position to form the output. Due to its simplicity and effectiveness, the nonlocal neural network\footnote{\scriptsize The nonlocal neural network is the deep CNNs with nonlocal block inserted} has been widely applied in image and video classification~\cite{nl, cgnl, ns, a2net}, image segmentation~\cite{ccnet, cgnl, nl} and person re-identification~\cite{nlreid, scan} tasks. 


The nonlocal (mean) operator in the nonlocal block\footnote{\scriptsize The nonlocal block consists a nonlocal operator and a residual connection} is related to the spatial-based graph convolution, which takes the aggregation of the central position/node and its neighbor positions/nodes in the image/graph to get a new representation. \citet{sgcn} shows the relation of the nonlocal block and the spatial-based graph convolutions in Message Passing Network~\cite{mpnn}. However, the nonlocal block focuses on a fully-connected graph, considering all the other positions in the image. Instead, the spatial based graph convolution focuses on a sparse graph, which naturally eliminates the redundant information. Specifically, the affinity matrix (the similarity metric of pairwise positions) computed in the nonlocal block is obtained from all the features of the upper layer, thus leading to the interference of feature aggregations. Therefore, the current nonlocal block needs an elaborate arrangement for its position, number, and channel number to eliminate this negative effect.

To increase the robustness and applicability of the nonlocal block in real-world applications, from the spectral-based graph convolution views~\cite{gcn, cheb_net, cay_net}, we reformulate the nonlocal block based on the property of graph spectral domain. As shown in Fig.~\ref{intro}, the input image is fed into the convolutional layers to extract discriminative features such as the wing, the head, the claw and the neck. These features can be seen as the input of the nonlocal block. Different from the spatial view which firstly aggregates the input features by weighted mean and then uses convolutional operator to filter as in Fig.~\ref{intro} \textbf{\sf{A}}, our spectral view constructs a fully-connected graph based on their similarity and then directly filters the input features in a global view profited by the graph filter shown in Fig.~\ref{intro} \textbf{\sf{B}}.



In practice, the Chebyshev polynomials is utilized to approximate the graph filter for reducing the number of parameters and computational cost~\cite{chebApr}. This approximated formulation has successfully filled the gap between the spectral view and the spatial view of the nonlocal block~\cite{survey_gcn}. Thus other extended nonlocal-based blocks ($\mathit{e.g.,}$ nonlocal block, nonlocal stage, double attention block) can be further theoretically interpreted in the spectral view.

Based on the points above, we propose Spectral Nonlocal Block (SNL) that concerns the rich, structured information in an image via encoding the graph structure. The SNL guarantees the existence of the graph spectral domain and has more mathematical guarantees to improve its robustness and accuracy. In a nutshell, our contributions are threefold:
\begin{itemize}
\item We have theoretically bridged the gap between nonlocal block (spatial-based approaches) and the graph spectral filter method (spectral-based approach)

\item We propose a spectral nonlocal block as an efficient, simple, and generic component of deep neural networks, which captures long-range spatial-temporal dependencies between spatial pixels or temporal frames.

\item The SNL achieves a clear-cut improvement over existing nonlocal blocks in multiple vision tasks, including image classification, fine-grained image classification, action recognition, and person re-identification.
\end{itemize}

\section{Preliminary}
In this paper, we use \textbf{bold} uppercase characters to denote the matrix-valued random variable and \textit{\textbf{italic bold}} uppercase to denote the matrix. Vectors are denoted with lowercase.

The Nonlocal Block (NL) follows the nonlocal operator that calculates a weighted mean between the features of each position and all possible positions as shown in Fig.~\ref{intro} \textbf{\sf{A}}. The nonlocal operator is defined as:
\begin{equation}
    F(\mX_{i, :}) = \frac{\sum_{j}\Big[f(\mX_{i, :}, \mX_{j, :})g(\mX_{j, :})\Big]}{\sum_{j} f(\mX_{i, :}, \mX_{j, :})}, 
\label{org_nlop}
\end{equation}
where $\mX \in \mathbb{R}^{N \times C_{1}}$ is the input feature map, $i, j$ are the position indexes in the feature map, $f(\cdot)$ is the affinity kernel which can adopt the ``Dot Product", ``Traditional Gaussian", ``Embedded Gaussian" or other kernel metrics with a finite Frobenius norm. $g(\cdot)$ is a linear embedding that is defined as: $g(\mX_{j, :}) = \mX_{j, :} \mathbf{W}_{Z}$ with $\mathbf{W}_{Z} \in \mathbb{R}^{C_{1} \times C_{s}}$. Here $N$ is the total positions of each features and $C_1, C_s$ are the number of channels for the input and the transferred features.
j
When inserting the NL block into the network structure, a linear transformation and a residual connection are added:
\begin{equation}
    \mY_{i, :} = \mX_{i, :} + F(\mX_{i, :}) \mathbf{W},
\label{org_nl}
\end{equation}
where $\mathbf{W} \in \mathbb{R}^{C_{s} \times C_{1}}$ is the weight matrix.

\section{Spectral Nonlocal (SNL) Block }
The nonlocal operator can be explained under the graph spectral domain. It can be briefly divided into two steps: generating a fully-connected graph to model the relation between the position pairs; converting the input features into the graph domain and learning a graph filter. In this section, we firstly give the definition of the spectral view for the nonlocal operator. Then, we interpret other nonlocal-based operators from this spectral view. Finally, we propose the Spectral Nonlocal (SNL) Block and highlight its properties.


\subsection{Nonlocal block in the spectral view} 
\label{sec_3_1}
The matrix form of the nonlocal operator in Eq.~(\ref{org_nlop}) is:
\begin{align}
F(\mX) = (\mD^{-1}_{M} \mM) \mZ \mathbf{W} = \mA \mZ \mathbf{W},
\label{nl_eq}
\end{align}
where $\mM = (M_{ij})$, $M_{ij} = f(\mX_{i,:}, \mX_{j,:})$ and $\mZ = \mX \mathbf{W}_{Z}$. The matrix $\mM \in \mathbb{R}^{N \times N}$ is composed by pairwise similarities between pixels. $\mZ \in \mathbb{R}^{N \times C_{s}}$ is the transferred feature map that compresses the channels of $\mX$ by a linear transformation with $\mathbf{W}_Z \in \mathbb{R}^{C_{1} \times C_{s}}$. $\mD_{M}$ is diagonal matrix which contains the degree of each vertex of $\mM$.

Then, we reformulate the nonlocal block by merging the nonlocal operator and the filter matrix $\mathbf{W}$:
\begin{equation}
    \mY = \mX + F(\mX)\mathbf{W} = \mX + \mathcal{F}(\mA, \mZ).
\label{eq:nl}
\end{equation}
$\mathcal{F}(\mA, \mZ$) is called as ``nonlocal operator" in the paper.

\textbf{Definition:} In this view, $\mathcal{F}(\mA, \mZ)$ can be seen as firstly computing the affinity matrix $\bm{A}$ that defines a graph spectral domain and then learns a filter for graph spectral features. Specifically, a fully-connected graph $\mathcal{G}=\{\mathbb{V}, \bm{A}, \bm{Z}\}$ is firstly constructed, in which $\mathbb{V}$ is the vertex set. Then, the node feature $\mZ$ is transformed into the graph spectral domain by the graph Fourier transformation $\mathscr{F}(\bm{Z})$. Finally, a graph filter $\mathbf{g}_{\theta}$ is generated to enhance the feature discrimination. Details of this graph Fourier transformation and the graph filter can be found in the Appendix.~\ref{appendix_gft}.

From this perspective, we interpret the nonlocal operator in the graph spectral view as below.
\begin{Theorem}
  Given an affinity matrix $\mA \in \mathbb{R}^{N \times N}$ and the signal $\mZ \in \mathbb{R}^{N \times C_{s}}$, the nonlocal operator is the same as filtering the signal $\mZ$ in the graph domain of a fully-connected weighted graph $\mathcal{G}$:
\begin{align}
        \mathcal{F}(\mA, \mZ) &= \mZ *_{\mathcal{G}} \mathbf{g}_{\theta} = \mU \mathbf{\Omega} \mU^\top \mZ     \label{eq:graphview}\\
        \quad \quad \mathrm{with} \quad \mL &= \mD_{L} - \mA = \mU^\top \bm{\Lambda} \mU \nonumber,
\end{align}
where the graph filter $\mathbf{\Omega} \in \mathbb{R}^{N \times N}$ is a diagonal matrix, i.e., $\mathbf{\Omega} = \mathrm{diag}(\mathbf{\omega} )$, $\mathbf{\omega} = (\omega_1, \omega_2, \cdots, \omega_n)$. The fully-connected graph $\mathcal{G} = (\mathbb{V}, \mZ, \mA)$ has the vertex set $\mathbb{V}$, node feature $\mZ$ and affinity matrix $\mA$. $\bm{\Lambda} = \mathrm{diag}\big(\{\lambda_1, \lambda_2, \cdots, \lambda_N\}\big)$ and $\mU = \{\mathbf{u}_1, \mathbf{u}_2, \cdots, \mathbf{u}_N\}$ are the eigenvalues and eigenvectors of the graph Laplacian $\mL$, respectively. $\mD_{L}$ is the degree matrix of $\mL$.
\label{T.1}
\end{Theorem}

The main difference between Theorem.~\ref{T.1} and the spatial view of nonlocal~\cite{nl} is that the former learns a graph filter to obtain the feature under the spectral domain while the latter filters the feature by the convolutional operator without concerning the graph structure.

\begin{Property} 
Theorem.\ref{T.1} requires the graph Laplacian $\mL$ has non-singular eigenvalues and eigenvectors. Thus, the affinity matrix $\mA$ should be non-negative and symmetric.
\label{R.1}
\end{Property}

\begin{Remark}
Based on Theorem.~\ref{T.1}, new nonlocal operators can be theoretically designed by using different types of graph filter such as the Chebyshev filter~\cite{cheb_net, gcn}, the graph wavelet filter~\cite{wavelets} and the Caylet filter~\cite{cay_net}. In this work, we utilize the Chebyshev filter.
\end{Remark}

\begin{Remark}
Different from the spectral-based graph convolutions that focus on the sparse graph with fixed structure~\cite{cheb_net}, Theorem.~\ref{T.1} focuses on the fully-connected graph structure in which affinity matrix $\mA$ is obtained by the affinity function on the deep features.
\end{Remark}

\begin{table*}[!htbp]
\centering
\small
\caption{Summary of nonlocal-based blocks in the spectral view. Our model has less constraints and is more flexible.}
\begin{tabular}{c|c|c|c|c|c}
\toprule
Models & Vertex ($|\mathbb{V}|$)& Edge ($|E|$)  & Affinity Matrix ($\mA$) & Node Feature ($\mZ$)& Formulation of ${F}(\mA, \mZ)$\\
\hline
\textbf{Spectral Form}  & - & - & - & - & $\mZ \mathbf{W}_{1} + \mA \mZ \mathbf{W}_{2} + \sum_{k=2}^{K-1} \mA^k \mZ \mathbf{W}_{k}$\\
\hline
NL & $N$ & $N \times N$ & $\mD^{-1}_{M} \mM$ & $\mX \mathbf{W}_Z$ & $\mA \mZ \mathbf{W}$ \\
\hline
$\mathrm{A^2}$ & $N$ & $N \times N$ & $\bf{\bar{M}}$ & $\mX \mathbf{W}_Z$ & $\mA \mZ \mathbf{W}$ \\
\hline
CGNL & $NC_s$ & $NC_s \times NC_s$ & $\mD^{-1}_{M} \mM$ & $\mathrm{vec}(\mX \mathbf{W}_Z)$ & $\mA \mZ \mathbf{W}$ \\
\hline
NS & $N$ & $N \times N$ &  $\mD^{-1}_{M} \mM$ & $\mX \mathbf{W}_Z$ & $ -\mZ \mathbf{W} + \mA \mZ \mathbf{W}$ \\
\hline
CC & $N$ & $N \times N$ & $\mD^{-1}_{\mC \odot \mM} (\mC \odot \mM)$ & $\mX$ & $\mA \mZ \mathbf{W}$ \\
\hline
Ours: \bf{SNL} & $N$ & $N \times N$ & $\mD^{-\frac{1}{2}}_{\hat{M}} \bm{\hat{M}} \mD^{-\frac{1}{2}}_{\hat{M}}$ & $\mX \mathbf{W}_Z$ & $ \mZ \mathbf{W}_{1} + \mA \mZ \mathbf{W}_{2}$ \\
\bottomrule
\end{tabular}
\label{relation}
\end{table*}

The graph filter $\mathbf{\Omega}$ in Eq.~(\ref{eq:graphview}) contains $N$ parameters. Due to the well-approximation and the conciseness~\cite{wavelets} of the Chebyshev polynomials, we utilize it to reduce the $N$ parameters into $k$ ($k$ is the order of polynomials, and $k \ll N$).
For simplicity, we assume that the input $\mZ$ has one channel. Then the graph filter approximated by $k_{\mathrm{th}}$-order Chebyshev polynomials is formulated as:
\begin{align}
  \mathcal{F}(\mA, \mZ) &= \sum_{k=0}^{K-1} \hat{\theta}_k T_k(\bm{\widetilde{L}}) \mZ,\label{eq:snl_1}\\
  \mathrm{s.t.} \quad
T_k({\bm{\widetilde{L}}}) &= 2 \bm{\widetilde{L}} T_{k-1}(\bm{\widetilde{L}}) - T_{k-2}(\bm{\widetilde{L}}),\nonumber
\end{align}
where $T_{0}(\bm{\widetilde{L}}) = \mI_{N}$ and $T_1({\bm{\widetilde{L}}}) = \bm{\widetilde{L}}$. $\hat{\theta}_{k}$ is the coefficient of the $k_{\mathrm{th}}$ term, $\bm{\widetilde{L}} = 2\mL/\lambda_{\mathrm{max}} - \mI_{N}$. Since $\mL$ is a normalized graph Laplacican, the maximum eigenvalue $\lambda_{\mathrm{max}} = 2$, which makes $\bm{\widetilde{L}} = -\mA$ \cite{gsp}. Thus Eq.~(\ref{eq:snl_1}) can be further derived as:
\begin{equation}
    \mathcal{F}(\mA, \mZ) = \theta_{0} \mZ + \theta_{1} \mA \mZ + \sum_{k=2}^{K-1} \theta_k \mA^{k} \mZ, 
\label{spectral_nonlocal_k}
\end{equation}
where $\theta_{k}$ can be learned via SGD.

Extending Eq.~(\ref{spectral_nonlocal_k}) into multiple channels, we can get a generalized formulation of the nonlocal operator:
\begin{equation}
\mathcal{F}(\mA, \mZ) = \mZ \mathbf{W}_{1} + \mA \mZ \mathbf{W}_{2} + \sum_{k=2}^{K-1} \mA^k \mZ \mathbf{W}_{k+1},
\label{full_SNL}
\end{equation}
where ${F}(\mA, \mZ)$ is the nonlocal operator, $\mathbf{W}_{k} \in \mathbb{R}^{C_{s} \times C_{1}}$.

\begin{Property}
Eq.~(\ref{full_SNL}) requires that the graph Laplacican $\mL$ is a normalized Laplacican whose maximum eigenvalue satisfies $\lambda_{\mathrm{max}} = 2$. Thus, $\bm{\widetilde{L}} = 2\mL/\lambda_{\mathrm{max}} - \mI_{N} = - \mA$. 
\label{R.L}
\end{Property}

\begin{Remark}
Eq.~(\ref{full_SNL}) gives the connection between spatial view and spectral view of the nonlocal operator, in which the graph filter is expressed by the aggregation between the $k_{th}$ neighbor nodes, i.e., all nodes for nonlocal. Thus, existing nonlocal-based structures can be theoretically analyzed by Eq.~(\ref{full_SNL}) in the spectral view.
\end{Remark}



\subsection{Relations with other nonlocal-based operators}
\label{sec_3_2}
Below we interpret other state-of-the-art nonlocal-based operators from the spectral view, and elaborate $5$ types of existing nonlocal-based blocks are special cases of Eq.~(\ref{full_SNL}) under certain graph structure and assumption. More detail proofs can be found in our Appendix.~\ref{appendix_prove_nl}.

\textbf{Original Nonlocal Block~\cite{nl}.}

Eq.~(\ref{nl_eq}) gives the formulation of the original nonlocal (NL) operator in the original nonlocal block. Our generalized formulation in Eq.~(\ref{full_SNL}) becomes the original NL operator when using the second term $\mA \mZ \mathbf{W}_{2}$ with the asymmetric affinity matrix $\mA = \mD^{-1}_{M} \mM$ to construct the graph.

\textbf{Nonlocal Stage~\cite{ns}.} 

To enable multiple NL operator in the deep networks, the Nonlocal Stage (NS) operator uses the graph Laplacian $\mL = \bm{I}_{N} - \mA$ as the affinity matrix $\mA$, making the NS follow the diffusion nature:
\begin{equation}
     \bar{\mathcal{F}}(\mA, \mZ) = (\mI_{N} - \mA) \mZ \mathbf{W},
\label{ns_eq}
\end{equation}
where $\bar{\mathcal{F}}(\cdot)$ is the NS operator. Our generalized formulation turns into the NS operator when using the $1_{st}$-order Chebyshev approximation $\mZ \mathbf{W}_{1} + \mA \mZ  \mathbf{W}_{2}$ with the assumption that $\mathbf{W}_{1} = -\mathbf{W}_{2}$.

\textbf{Double Attention Block~\cite{a2net}.}

To gather features in the entire space and then distribute them back to each location, \citet{a2net} proposes a two-step operator which can be reformulated as:
\begin{align}
    \mathcal{F}_{a}(\mX) 
    &= \Big[\sigma(\mX \mathbf{W}_{\phi}) \sigma(\mX \mathbf{W}_{\psi})\Big] (\mX \mathbf{W}_{g}) \mathbf{W} \nonumber\\
    &= \bm{\bar{M}} \mZ \mathbf{W} = \bm{\bar{A}} \mZ \mathbf{W} = \mathcal{F}_{a}( \bm{\bar{A}}, \mZ), 
\end{align}
where $\mathcal{F}_{a}(\cdot)$ is the ``$\mathrm{A^2}$ operator" and $\sigma(\cdot)$ is the softmax function. $\mathbf{W}_{\phi,\psi,g} \in \mathbb{R}^{C_{1} \times C_{s}}$ are three weight matrices. Our generalized formulation turns into the $\mathrm{A^2}$ operator when only using the second term $\bm{\bar{A}} \mZ \mathbf{W}_{2} $ with the affinity matrix $\bm{\bar{A}} = \sigma(\mX \mathbf{W}_{\phi}) \sigma(\mX \mathbf{W}_{\psi})$. Compared with the original NL, the two softmax operators used to form $\bm{\bar{A}}$ can eliminate fallacious value of $\mX \mathbf{W}_{\phi}, \mX \mathbf{W}_{\psi}$ (\textit{i.e.,} negative value).

\textbf{Compact Generalized Nonlocal Block ~\cite{cgnl}.}

The compact generalized nonlocal (CGNL) block catches cross-channel clues, which generalizes the nonlocal block via further considering the correlations between pairwise channels. The CGNL operator can be reformulated as:
\begin{align}
    \mathcal{F}_{g}(\mA, \mZ) &= \mA \bm{z} W = \mD_M^{-1}\mM \bm{z} W, \label{cgnl_eq}\\
    \mathrm{with} \quad \bm{z} &= \mathrm{vec}(\mZ) \in \mathbb{R}^{NC \times 1}, \nonumber
\end{align}
where $\mathcal{F}_{g}(\cdot)$ is the CGNL operator, $\mathrm{vec}(\mZ)$ reshapes the feature $\mZ$ by merging channel into positions. Thus, when constructing a more complex fully-connected graph that has $N \times C_s$ nodes and using the second term for approximation, our generalized formulation turns into the CGNL operator. Although the CGNL operator can model the dependencies between channels by its complex graph structure, it also increases the parameters of graph filter from $N^2C_s$ to $N^2C_s^2$.

\textbf{Criss-Cross Attention Block~\cite{ccnet}}

To model long-range dependencies with the lightweight architecture and efficient computation, \citet{ccnet} introduces the criss-cross attention block (CC), which can be further reformulated as:
\begin{align}
     \mathcal{F}_{c}(\mX) = (\mC \odot (\mD_M^{-1}\mM)) \mZ \mathbf{W} =  \widetilde{\bm{A}} \mX \mathbf{W}\\
     \mathrm{with} \quad
       \mC_{ij} =  
        \begin{cases}
        1& r_i = r_j || c_i = c_j\\
        0& \text{else}
        \end{cases} , \nonumber
\end{align}
where $\mathcal{F}_{c}(\cdot)$ is the CC operator, $r_i$ is the index of row, $c_i$ is the index of column and $\odot$ is the Hadamard product. We can see that our generalized formulation can also turn into the CC operator when using the second term $ \bm{\widetilde{A}} \mX \mathbf{W}_{2}$ with the node feature $\mX$. In this view, the CC operator can be seen as masking some edges of the fully-connected graph structure (by setting $\mC_{ij} = 0$) to improve the efficiency.

\begin{figure*}[!htbp]
    \centering
    \includegraphics[width=.9\textwidth]{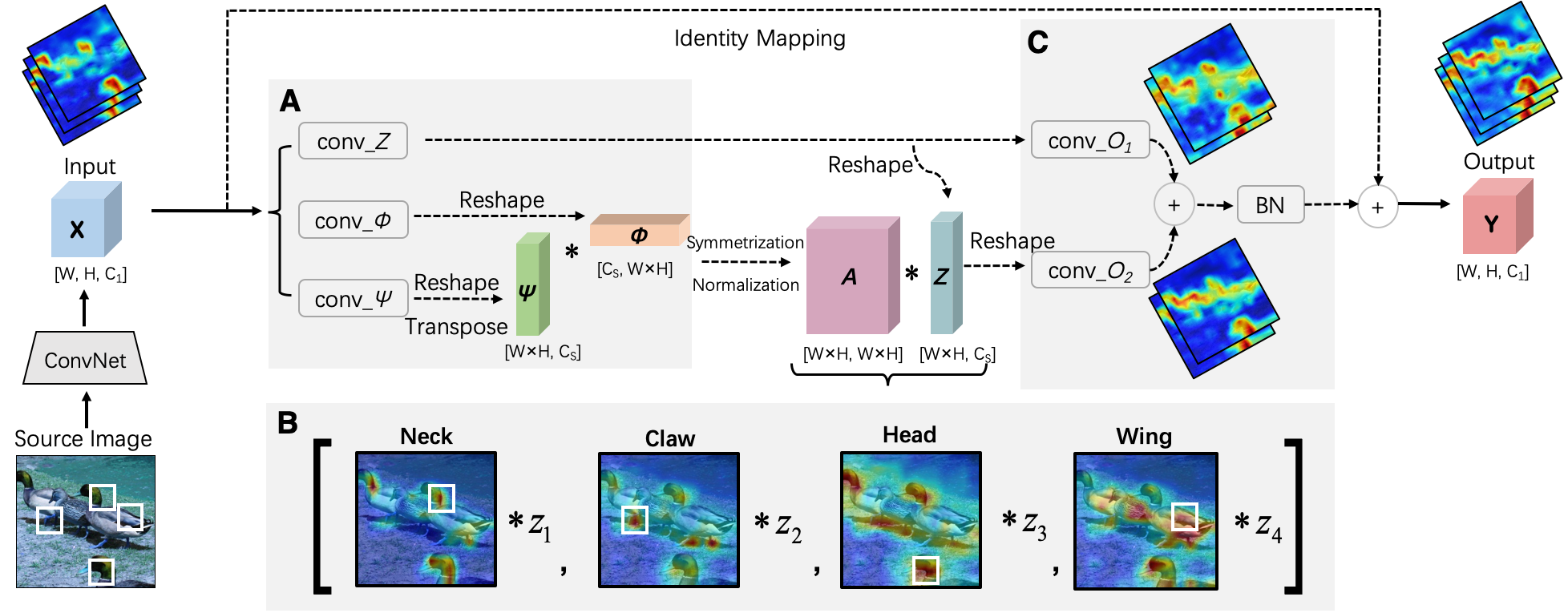}
    \caption{The implementation of the Spectral Nonlocal Block. \textbf{ \sf{A}.} Three feature maps $\bm{\phi}$, $\bm{\psi}$, $\mZ$ are generated by feeding the input feature map $\tX$ into three $1\times1$ convolutions. Then, a normalized symmetrization affinity matrix $\mA$ is obtained by generating affinity kernel on $\bm{\phi}$ and $\bm{\psi}$. \textbf{ \sf{B}.} The second term of Eq.~(\ref{snl}) is calculated with $\mZ$ and the affinity matrix $\mA$. Each row of $\mA$ contains a $N$-dimension spatial attention map (heat maps) and $\mathbf{z}_{1}, \mathbf{z}_{2}, \cdot\cdot\cdot, \mathbf{z}_{n}$ are the column vectors of $\mZ$ (for simplicity, here we pick $n=4$ where the white squares are the central positions we visualize). \textbf{\sf{C}.} The graph filter is approximated by respectively feeding $\mA \mZ$, $\mZ$ into two convolutions and then add them together as in Eq.~(\ref{snl}). Finally, batch normalization is conducted and the result is added with $\tX$ to obtain the output $\tY$.}\vspace{-2mm}
    \label{gnl_block}
\end{figure*}

\subsection{The proposed spectral nonlocal block}
\label{sec_3_3}

\textbf{Formulation:}
Based on Eq.~(\ref{full_SNL}), we propose a more rational nonlocal-based block called the Spectral Nonlocal Block (SNL) that concerns the property under the graph spectral domain with more complete approximation as:
\begin{align}
    &\bm{Y} = \mX + \mathcal{F}_s(\mA, \mZ) = \mX + \mZ \mathbf{W}_{1} + \mA \mZ \mathbf{W}_{2}, \label{snl}\\
    &\mathrm{s.t.}  \quad \mA = \mD^{-\frac{1}{2}}_{\hat{M}} \bm{\hat{M}} \mD^{-\frac{1}{2}}_{\hat{M}} ,\quad \bm{\hat{M}} = (\mM + \mM^{\top}) / 2 \nonumber
\end{align}
where $\mathcal{F}_{s}(\mA, \mZ)$ is the SNL operator, $\mathbf{W}_{1},\mathbf{W}_{2} \in \mathbb{R}^{C_{s} \times C_{1}}$, $\mathbf{W}_{2}$ are two parameter matrixes.

\begin{Remark}
The proposed SNL uses a symmetric affinity matrix $\mA = \mD^{-\frac{1}{2}}_{\hat{M}} \bm{\hat{M}} \mD^{-\frac{1}{2}}_{\hat{M}}$ to ensure the existence of the real eigenvalue, which definitely satisfies the precondition of defining a graph filter. Thus, our SNL is more stable when inserted into the deep neural networks.
\end{Remark}

\begin{Remark}
The proposed SNL uses the complete form of $1_{st}$-order Chebyshev Approximation which is a more accurate approximation of the graph filter. Thus, our SNL can give the parameters a liberal learning space with only one more parameter matrix.
\end{Remark}

\begin{table*}[!htbp]
\centering
\small
\caption{The Performances of Nonlocal-based Blocks with Different Number of Transferred Channels on CIFAR-100}
\label{Tab:subchannel}
\renewcommand\tabcolsep{3.0pt}
\begin{tabular}{lcc|lcc|lcc}
\toprule
\multicolumn{3}{c|}{No Reduction} & \multicolumn{3}{c|}{Reduction by Two Times} & \multicolumn{3}{c}{Reduction by Four Times}\\
\hline
Models & Top1 (\%) & Top5 (\%) & Models & Top1 (\%) & Top5 (\%)  & Models & Top1 (\%) & Top5 (\%) \\
\hline
PreResNet56 & $ 75.33^{\uparrow0.00}$ & $ 93.97^{\uparrow0.00}$ & 
PreResNet56 & $ 75.33^{\uparrow0.00}$ & $ 93.97^{\uparrow0.00}$ &
PreResNet56 & $ 75.33^{\uparrow0.00}$ & $ 93.97^{\uparrow0.00}$ \\

+ NL & $ 75.29^{\downarrow0.04}$ & $ 94.07^{\uparrow0.10}$ & 
+ NL & $ 75.31^{\downarrow0.02}$ & $ 92.84^{\downarrow1.13}$ &
+ NL & $ 75.50^{\uparrow0.17}$ & $ 93.75^{\downarrow0.22}$ \\

+ NS & $ 75.39^{\uparrow0.06}$ & $ 93.00^{\downarrow0.97}$ & 
+ NS & $ 75.83^{\uparrow0.50}$ & $ 93.87^{\downarrow0.10}$ &
+ NS & $ 75.61^{\uparrow0.28}$ & $ 93.66^{\downarrow0.31}$ \\

+ $\mathrm{A^2}$  & $ 75.51^{\uparrow0.18}$ & $ 92.90^{\downarrow1.07}$ & 
+ $\mathrm{A^2}$  & $ 75.58^{\uparrow0.25}$ & $ 94.27^{\uparrow0.30}$ &
+ $\mathrm{A^2}$   & $ 75.61^{\uparrow0.28}$ & $ 93.61^{\downarrow0.36}$ \\

+ CGNL & $ 74.71^{\downarrow0.62}$ & $ 93.60^{\downarrow0.37}$ & 
+ CGNL & $ 75.75^{\uparrow0.42}$ & $ 93.74^{\downarrow0.23}$ &
+ CGNL & $ 75.27^{\downarrow0.06}$ & $ 93.05^{\downarrow0.92}$ \\

\hline\hline
\textbf{+ SNL} & $ \mathbf{\underline{76.34}^{\uparrow1.01}}$ & $ \mathbf{\underline{94.48}^{\uparrow0.51}}$ & 
\textbf{+ SNL} & $ \mathbf{\underline{76.41}^{\uparrow1.08}}$ & $ \mathbf{\underline{94.38}^{\uparrow0.41}}$ &
\textbf{+ SNL} & $ \mathbf{\underline{76.02}^{\uparrow0.69}}$ & $ \mathbf{\underline{94.08}^{\uparrow0.11}}$ \\
\bottomrule
\vspace{-5mm}
\end{tabular}
\end{table*}


\begin{table*}[!htbp]
\centering
\small
\renewcommand\tabcolsep{3.0pt}
\caption{The Performances of Nonlocal-based Blocks Inserted into Different Position of Deep Networks on CIFAR-100}
\label{Tab:stage}
\begin{tabular}{lcc|lcc|lcc}
\toprule
\multicolumn{3}{c|}{Stage 1} & \multicolumn{3}{c|}{Stage 2} & \multicolumn{3}{c}{Stage 3}\\
\hline
Models & Top1 (\%) & Top5 (\%) & Models & Top1 (\%) & Top5 (\%)  & Models & Top1 (\%) & Top5 (\%) \\
\hline

PreResNet56 & $ 75.33^{\uparrow0.00}$ & $ 93.97^{\uparrow0.00}$ & 
PreResNet56 & $ 75.33^{\uparrow0.00}$ & $ 93.97^{\uparrow0.00}$ &
PreResNet56 & $ 75.33^{\uparrow0.00}$ & $ \mathbf{\underline{93.97}^{\uparrow0.00}}$ \\

+ NL & $ 75.31^{\downarrow0.02}$ & $ 92.84^{\downarrow1.13}$ &
+ NL & $ 75.64^{\uparrow0.31}$ & $ 93.79^{\downarrow0.18}$ & 
+ NL & $ 75.28^{\downarrow0.05}$ & $ 93.93^{\downarrow0.04}$ \\

+ NS & $ 75.83^{\uparrow0.50}$ & $ 93.87^{\downarrow0.10}$ &
+ NS & $ 75.74^{\uparrow0.41}$ & $ 94.02^{\uparrow0.05}$ &
+ NS & $ 75.44^{\uparrow0.11}$ & $ 93.86^{\downarrow0.11}$ \\

+ $\mathrm{A^2}$  & $ 75.58^{\uparrow0.25}$ & $ 94.27^{\uparrow0.30}$ &
+ $\mathrm{A^2}$  & $ 75.60^{\uparrow0.27}$ & $ 93.82^{\downarrow0.15}$ &
+ $\mathrm{A^2}$   & $ 75.21^{\downarrow0.12}$ & $ 93.65^{\downarrow0.32}$ \\

+ CGNL & $ 75.75^{\uparrow0.42}$ & $ 93.74^{\downarrow0.23}$ &
+ CGNL & $ 74.54^{\downarrow0.79}$ & $ 92.65^{\downarrow1.32}$ &
+ CGNL & $ 74.90^{\downarrow0.43}$ & $ 92.46^{\downarrow1.51}$ \\

\hline\hline
\textbf{+ SNL} & $ \mathbf{\underline{76.41}^{\uparrow1.08}}$ & $ \mathbf{\underline{94.38}^{\uparrow0.41}}$ &
\textbf{+ SNL} & $ \mathbf{\underline{76.29}^{\uparrow0.96}}$ & $ \mathbf{\underline{94.27}^{\uparrow0.30}}$ &
\textbf{+ SNL} & $ \mathbf{\underline{75.68}^{\uparrow0.35}}$ & $ 93.90^{\downarrow0.07}$ \\
\bottomrule
\end{tabular}
\end{table*}

The summary of our SNL and other nonlocal-based operators in the spectral view is shown in Table~\ref{relation}. For the affinity matrix, we can see except for our SNL, other nonlocal-based operators use the random walk normalized (NL, NS, CGNL, CC) or the non-normalized affinity matrix ($\mathrm{A^2}$) whose symmetry is not guaranteed and depends on the affinity kernel. This makes the affinity matrix against Property.~\ref{R.1} and leads to the non-existence of the graph spectral domain. Thus, their robustness and flexibility are weakened. 

For the approximation, we can see that except for our SNL, all other nonlocal-based operators only use the second term (NL, $\mathrm{A^2}$, CGNL, CC) or the $1_{st}$-order approximation with sharing weight (NS) rather than the complete form of the $1_{st}$-order approximation, which hinders their performance. However, our SNL uses a symmetry affinity matrix and a more complete approximation, which improves the robustness and performance when added into the deep neural network.


\textbf{Implementation:}
The implementation details of the SNL block is shown in Fig.~\ref{gnl_block}. The input feature map $\tX \in \mathbb{R}^{W \times H \times C_1}$ is firstly fed into three 1$\times$1 convolutions with the weight kernels: $\mathbf{W}_{\phi,\psi,g} \in \mathbb{R}^{C_{1} \times C_{s}}$ to subtract the number of channels and then reshaped into $\mathbb{R}^{WH \times C_{s}}$. One of the output $\mZ \in \mathbb{R}^{WH \times C_{s}}$ is used as the transferred feature map to reduce the calculation complexity, while the other two outputs $\bm{\varPhi},\bm{\varPsi} \in \mathbb{R}^{WH \times C_{s}}$ are used to get the affinity matrix $\mA$ with the affinity kernel function $f(\cdot)$. Then, $\mA$ is made to be symmetric and normalized as in Eq.~(\ref{snl}). Finally, with the affinity matrix $\mA$ and the transferred feature map $\mZ$, the output of the nonlocal block can be obtained by the Eq.~(\ref{snl}). Specifically, the two weight matrices $\mathbf{W}_{1,2} \in \mathbb{R}^{C_{s} \times C_{1}}$ are implemented by two 1$\times$1 convolutions.

\section{Experiments}
In this section, we design the ablation experiments to test the robustness of nonlocal-based blocks with different numbers, different positions, and different channels when inserted into deep models. Then, we show performance of the proposed SNL in $4$ vision tasks, including image classification (Cifar-10/100~\footnote{\scriptsize https://www.cs.toronto.edu/~kriz/cifar.html}, ImageNet~\footnote{\scriptsize http://www.image-net.org/}), fine-grained image classification (CUB-200~\footnote{\scriptsize http://www.vision.caltech.edu/visipedia/CUB-200.html}), action recognition (UCF-101~\footnote{\scriptsize https://www.crcv.ucf.edu/data/UCF101.php}). The experimental results on the person re-identification tested with ILID-SVID~\cite{ilid_reid}, Mars~\cite{mars_reid}, and Prid-2011~\cite{prid_reid} datasets are given in the Appendix.~\ref{appendix_reid}. All the methods are implemented using PyTorch~\cite{pytorch} toolbox with an Intel Core i9 CPU and $8$ Nvidia RTX 2080 Ti GPUs.

\subsection{Ablation experiments on CIFAR-100}
\label{sec_4_1}
\textbf{Experimental Setup}
The ablation experiments are conducted on CIFAR-100 dataset which contains $60,000$ images of $100$ classes. We use $50,000$ images as the training set and $10,000$ images as the testing set. PreResNet56~\cite{preresnet} is used as the backbone network. Unless otherwise specified, we set $C_s = C_1 / 2$ and add $1$ nonlocal-based block right after the second residual block in the early stage ($res1$). The initial learning rate $0.1$ is used with the weight decay $10^{-4}$ and momentum  $0.9$. The learning rate is divided by $10$ at $150$ and $250$ epochs. All the models are trained for $300$ epochs. We choose the evaluation criterion of the classification accuracy: Top1 and Top5 accuracy, which means the model prediction (the one with the highest probability) is exactly the expected label and 5 highest probability predictions contains the expected label. 

\textbf{The number of channels in transferred feature space}
The nonlocal-based block firstly reduces the channels of original feature map $C_{1}$ into the transferred feature space $C_{s}$ to reduce the computation complexity. If $C_{s}$ is too large, the feature map will contain redundant information which introduces the noise when calculating the affinity matrix $\mA$. However, if $C_{s}$ is too small, it is hard to reconstruct the output feature map due to inadequate features. To test the robustness for the value of the $C_{s}$, we generate three types of models with different $C_{s}$ setting: ``No Reduction" ($C_{s} = C_{1}$), ``Reduction by Two Times" ($C_{s} = C_{1}/2$), ``Reduction by Four Times " ($C_{s} = C_{1} / 4$). Table~\ref{Tab:subchannel} shows the experimental results of the $3$ types of models with different nonlocal-based blocks. Our SNL block outperforms other models profited by the flexibility for learning. 

Moreover, from Table~\ref{Tab:subchannel}, we can see that the performances of the CGNL steeply drop when the number of the transferred channels increases. This is because the CGNL block concerns the relations between channels. When the number of the transferred channels increases, the relations between the redundant channels seriously interfere with its effects. Overall, our SNL block is the most robust for the large number of transferred channels (our model rises $1.01\%$ in Top1 while the best of others only rise $0.18\%$ over the backbone).

\begin{table*}[!htbp]
\centering
\small
\caption{Experiments for Adding Different Types of Nonocal-based Blocks into PreResnet56 and ResNet50 on CIFAR-10/100 Dataset}
\label{tab:cifar}
\begin{tabular}{lcc|lcc|lcc}
\toprule
 \multicolumn{3}{c}{CIFAR-10} & \multicolumn{6}{|c}{CIFAR-100}\\
\hline
Models & Top1 (\%) & Top5 (\%) & Models & Top1 (\%) & Top5 (\%)  & Models & Top1 (\%) & Top5 (\%) \\
\hline
ResNet50 & $94.94^{\uparrow0.00}$ & $99.87^{\uparrow0.00}$ &
PreResnet56 & $75.33^{\uparrow0.00}$ & $93.97^{\uparrow0.00}$ & 
ResNet50 & $76.50^{\uparrow0.00}$ & $93.14^{\uparrow0.00}$\\

+ NL & $94.01^{\downarrow0.93}$ & $99.82^{\downarrow0.05}$ &
+ NL & $75.31^{\downarrow0.02}$ & $92.84^{\downarrow1.33}$ &
+ NL & $76.77^{\uparrow0.27}$ & $93.55^{\uparrow0.41}$ \\

+ NS & $95.15^{\uparrow0.21}$ & $99.88^{\uparrow0.01}$ & 
+ NS & $75.83^{\uparrow0.50}$ & $93.87^{\downarrow0.10}$ & 
+ NS & $77.90^{\uparrow1.40}$ & $\mathbf{\underline{94.34}^{\uparrow1.20}}$ \\

+ $\mathrm{A^2}$   & $94.41^{\downarrow0.53}$ & $99.83^{\downarrow0.05}$ &
+ $\mathrm{A^2}$  & $75.58^{\uparrow0.25}$ & $94.27^{\uparrow0.30}$ & 
+ $\mathrm{A^2}$   & $77.30^{\uparrow0.80}$ & $93.40^{\uparrow0.26}$ \\

+ CGNL & $94.49^{\downarrow0.45}$ & $99.92^{\uparrow0.05}$ &
+ CGNL & $75.75^{\uparrow0.42}$ & $93.74^{\downarrow0.23}$ &
+ CGNL & $74.88^{\downarrow1.62}$ & $92.56^{\downarrow0.58}$\\

\hline\hline
\textbf{+ SNL} & $\mathbf{\underline{95.32}^{\uparrow0.38}}$ & $\mathbf{\underline{99.94}^{\uparrow0.07}}$ &
\textbf{+ SNL} & $\mathbf{\underline{76.41}^{\uparrow1.08}}$ & $\mathbf{\underline{94.38}^{\uparrow0.39}}$ &
\textbf{+ SNL} & $\mathbf{\underline{78.17}^{\uparrow1.67}}$ & $94.17^{\uparrow1.03}$\\
\bottomrule
\vspace{-6mm}
\end{tabular}
\end{table*}

\textbf{The stage/position for adding the nonlocal-based blocks}
The nonlocal-based blocks can be added into the different stages of the preResNet to form the Nonlocal Network. In \citet{ns}, the nonlocal-based blocks are added into the early stage of the preResNet to catch the long-range relations. Here we show the performances of adding different types of nonlocal-based blocks into the $3$ stages (the first, the second and the third stage of the preResNet). The experimental results are shown in Table~\ref{Tab:stage}. We can see that the performances of the NL block is lower than the backbones when added into the early stage. However, our proposed SNL block has averagely $1.08\%$ improvement over the backbone when added into the early stage, which is more than twice over other types of nonlocal-based blocks ($0.42\%$ for the best case).

\textbf{The number of the nonlocal-based blocks} We test the robustness for adding nonlocal-based blocks into the backbone. The results are shown in Table~\ref{tab:number}. ``$\times 3$" means three blocks are added into the stage $1,2$, and $3$ respectively, and the accuracy in the brackets represent their results. We can see that adding three proposed SNL operators into different stages of the backbone generates a larger improvement (1.37\%) than the NS operator and NL operator. This is because when adding NS and NL into the early stage, these two models cannot well aggregate the low-level features and interfere with the following blocks. 
\begin{table}[!htbp]
\centering
\small
\vspace{-3mm}
\caption{Experiments for Adding Different Number of Nonlocal-based Blocks into PreResNet56 on CIFAR-100}
\renewcommand\tabcolsep{1.7pt}
\label{tab:number}
\begin{tabular}{lc|c}
\toprule
Models & Top1 (\%) & Top5 (\%)\\
\hline
PreResNet56 & $ 75.33^{\uparrow0.00}$ & $ 93.97^{\uparrow0.00}$\\
+ NL ($\times 3\dagger$)& $ 75.31^{\downarrow0.02}$ ($74.34^{\downarrow0.99}$) & $ 92.84^{\downarrow1.13}$ ($ 93.11^{\downarrow0.86}$)\\

+ NS ($\times 3$)& $ 75.83^{\uparrow0.50}$ ($75.00^{\downarrow0.33}$) & $ 93.87^{\downarrow0.10}$ ($ 93.57^{\downarrow0.40}$)\\

+ $\mathrm{A^2}$ ($\times 3$) & $ 75.58^{\uparrow0.25}$ ($ 75.63^{\uparrow0.33}$) & $ 94.27^{\uparrow0.30}$ ($\mathbf{\underline{94.12}^{\uparrow0.15}}$) \\

+ CGNL  ($\times 3$) & $ 75.75^{\uparrow0.42}$ ( $75.96^{\uparrow0.63}$) & $ 93.74^{\downarrow0.23}$ ($ 93.10^{\downarrow0.87}$)\\

\hline\hline
\textbf{+ SNL}  ($\times 3$) & $\mathbf{\underline{76.41}^{\uparrow1.08}}$  ($ \mathbf{\underline{76.70}^{\uparrow1.37}}$)& 
$\mathbf{\underline{94.38}^{\uparrow0.41}}$ ($93.94^{\downarrow0.03}$) \\
\bottomrule
\end{tabular}
\begin{flushleft}
$\dagger$ \textit{The number in the bracket means the performance when adding $3$ this kind of nonlocal-based blocks in the network.}
\end{flushleft}
\vspace{-5mm}

\end{table}



\subsection{Applications on computer vision tasks}
\label{sec_4_2}

\textbf{Image Classification}
CIFAR-10/100 and ImageNet are tested and our SNL outperforms other types of the nonlocal-based blocks on these standard benchmarks. We use the ResNet50~\cite{resnet} as the backbone and insert the SNL block right before the last residual block of \textit{res4} for fair comparison. Other settings for the CIFAR-10/100 are the same as our ablation experiments (Sec.~\ref{sec_4_1}). For the ImageNet, the initial learning rate $0.01$ is used with the weight decay $10^{-4}$ and momentum  $0.9$. The learning rate is divided by $31$ at $61$ and $81$ epochs. All the models are trained for $110$ epochs. The ``floating-point operations per second" (Flops) and the ``model size" (Size) are used to compare the computation complexity and memory consumption.
 
Table~\ref{tab:cifar} shows the experimental results on the CIFAR10 dataset. When adding one proposed block, the Top1 classification accuracy rises about $0.38\%$, which is nearly twice over other types of nonlocal-based blocks (the best is $0.21\%$). As the experiments on CIFAR100 shown in Table~\ref{tab:cifar}, using our proposed block brings significant improvements about $1.67\%$ with ResNet50. While using a more simple backbone PreResnet56 as shown in Table~\ref{tab:cifar}, our model can still generate $1.08\%$ improvement which is not marginal.

The results of ImageNet are shown in Table~\ref{tab:imagenet}. 
Note that other baselines are reported with the scores in their paper. We can see that compared with the nonlocal-based blocks, our SNL achieves a clear-cut improvement ($1.96\%$) with a minor increment in complexity ($0.51$G Flops and $2.62$M of Size compared with original $4.14$G Flops and $25.56$M). Moreover, our SNL is also superior to other types of blocks such as SE block~\cite{senet}, CGD block\cite{cgdnet}, GE block\cite{genet} ($0.11$\% higher in Top1 and $2.02$M lower in size than the GE block).

\begin{table}[!htbp]
\centering
\small
\vspace{-3mm}
\caption{Experiments for adding different types of nonocal-based blocks into Resnet50 on ImageNet}
\label{tab:imagenet}
\begin{tabular}{lccc}
\toprule
Models & Top1 (\%) & Flops (G) & Size (M)\\
\hline
ResNet50 & $76.15^{\uparrow0.00}$ & $4.14$ & $25.56$ \\ 
+ CGD & $76.90^{\uparrow0.75}$ & $+ 0.01$ & $+ 0.02$ \\ 
+ SE & $77.72^{\uparrow1.57}$ & $+ 0.10$ & $+ 2.62$\\
+ GE & $78.00^{\uparrow1.85}$ & $+0.10$ & $+5.64$ \\
\cline{2-4}
+ NL & $76.70^{\uparrow0.55}$ & $+ 0.41$ & $+ 2.09$\\
+ $\mathrm{A^2}$ & $77.00^{\uparrow0.85}$  & $+ 0.41$ & $+ 2.62$\\
+ CGNL & $77.32^{\uparrow1.17}$ & $+ 0.41$ & $+ 2.09$\\
\hline\hline
 \textbf{+ SNL} & $\mathbf{\underline{78.11}^{\uparrow1.96}}$ & $+0.51$ & $+ 2.62$\\
\bottomrule
\end{tabular}
\end{table}

\begin{figure*}[!htbp]
    \centering
    \includegraphics[width=0.95\textwidth]{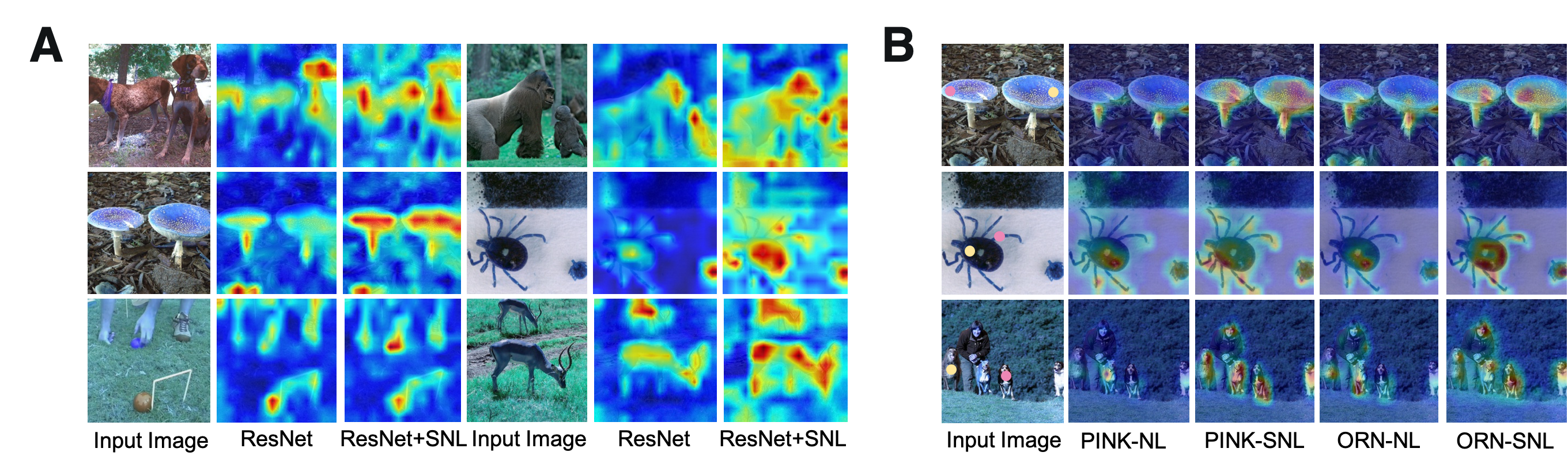} \vspace{-3mm}
    \caption{
    \textbf{ \sf{A}.} The visualization of the feature maps when adding SNL into the ResNet50 backbone.
    \textbf{ \sf{B}.} The visualization of the attention maps for two positions (``Pink" and ``Orange" dots). The heatmaps show the strength of similarity between them and other positions.
    }\vspace{-2mm}
    \label{fig:imagenet_feature_attention}
\end{figure*}


We also visualize the output feature maps of the ResNet50 with SNL and the original ResNet50 in Fig.~\ref{fig:imagenet_feature_attention} \textbf{\sf{A}}. Benefited from the rich and structured information considered in SNL, the response of the similar features between long-range spatial positions are enhanced as shown in the two mushroom, balls, and those animals. Moreover, Fig.~\ref{fig:imagenet_feature_attention} \textbf{\sf{B}} shows the attention maps produced by our SNL and the original NL block where the ``Pink" and ``Orange" dots are the central positions and the heatmaps represent the similarity between the central position and other positions. Compared with the original NL block, SNL can pay more attention to the crucial parts than the original NL block profited by the better approximation formation as discussed in Sec.~\ref{sec_3_3}.



\textbf{Fine-grained Image Classification}
The experiments for the fine-grained classification are generated on the Birds-200-2011 (CUB-200) dataset which contains $11,788$ images of $200$ categories of different birds. We use $5,994$ images as the training set and $5,794$ images as the testing set as \citet{cgnl}. We use the ResNet50 model pre-trained on ImageNet as the backbone and train the models for total $110$ epochs with the initial learning rate $0.1$ which is subsequently divided by $10$ at $31$, $61$, $81$ epochs. Table~\ref{tab:cub_ucf} (CUB-200) shows that our model can generate ($0.59\%$) improvement. Compared with the CGNL block concerning channel-wise relations, our SNL is just a bit lower in Top1 ($0.12\%$). That is because the dependencies between channels play an important role in the fine-grained classification. However, these channel dependencies of CGNL can impede the practical implementations, which needs elaborate preparations for the number of channels per block, the number of blocks and their positions as shown in Table~\ref{Tab:subchannel}, \ref{Tab:stage}, \ref{tab:number}. Compared with the other non-channel concerned nonlocal block, our SNL has improvements with a large margin.

\begin{table}[!htbp]
\centering
\small
\vspace{-2mm}
\caption{Experiments for Nonocal-based Blocks Added into ResNet50 and I3D on CUB-200 and UCF-101 Datasets}
\label{tab:cub_ucf}
\renewcommand\tabcolsep{1.7pt}
\begin{tabular}{l|cc|cc}
\toprule
& \multicolumn{2}{c|}{CUB-200} & \multicolumn{2}{c}{UCF-101}\\
\hline
Models & Top1 (\%) & Top5 (\%) & Top1 (\%) & Top5 (\%) \\
\hline
ResNet50/I3D$^\dagger$  &
 $85.43^{\uparrow0.00}$ & $96.70^{\uparrow0.00}$ & $81.57^{\uparrow0.00}$ & $95.40^{\uparrow0.00}$\\

+ NL & $85.34^{\downarrow0.09}$ & $\mathbf{\underline{96.77}^{\uparrow0.07}}$  &  $82.88^{\uparrow1.31}$ & $95.74^{\uparrow0.34}$  \\

+ NS  & $85.54^{\uparrow0.11}$ & $96.56^{\downarrow0.14}$ & $82.50^{\uparrow0.93}$ &  $95.84^{\uparrow0.44}$\\

+ $\mathrm{A^2}$   
& $85.91^{\uparrow0.48}$ & $96.56^{\downarrow0.14}$ & $82.68^{\uparrow1.11}$ &  $95.85^{\uparrow0.45}$ \\

+ CGNL  &
\textbf{$\mathbf{\underline{86.14}^{\uparrow0.71}}$} & $96.34^{\downarrow0.36}$ & $83.38^{\uparrow1.81}$ & $95.42^{\uparrow0.02}$\\

\hline\hline
\textbf{+ SNL} &
$\mathbf{\underline{86.02}^{\uparrow0.59}}$ & $96.65^{\downarrow0.05}$ & $\mathbf{\underline{84.39}^{\uparrow2.82}}$ & $\mathbf{\underline{97.66}^{\uparrow2.26}}$  \\
\bottomrule
\end{tabular}
\begin{flushleft}
$\dagger$ \textit{The ResNet50 is used for CUB-200 dataset as the network bones and I3D is used for UCF-101 dataset.}
\end{flushleft}
\end{table}

\textbf{Action Recognition}
Experiments are conducted on the UCF-101 dataset, which contains $9,537$ videos for $101$ different human actions. We use $7,912$ videos as the training set and $1,625$ videos as the testing set. Our SNL block are tested on the UCF-101 dataset for capturing the dependence for the temporal frames. We follow the I3D structure~\cite{I3d} which uses $k \times k \times k$ kernels to replace the convolution operator in the residual block for learning seamless spatial-temporal feature extractors. The weights are initialized by the pre-trained I3D model on Kinetics dataset \cite{kinetics}. Inserting nonlocal-based blocks into the I3D can help to capture the relations between frame pairs with long distance and improve the feature representation. We train the models with the initial learning rate of $0.01$ which is subsequently divided by $10$ each $40$ epochs. The training stops at the $100$ epochs. Other hyper-parameters of the experimental setup are the same as in Sec.~\ref{sec_4_1}. 

Table~\ref{tab:cub_ucf} (UCF-101) shows the results on the action recognition. The network with our proposed block can generate significant improvements ($2.82\%$) than the I3D and outperforms all other nonlocal-based models on the UCF-101 dataset. This shows that our proposed SNL is also effective for catching the long-range dependencies between the temporal frames. We also conduct the experiments on UCF-101 dataset with other state-of-the-art action recognition models in our Appendix.~\ref{appendix_stoa} including the P3D~\cite{p3d}, the MARS~\cite{mars}, and the VTN~\cite{vtn}. All of them demonstrate that our SNL block can easily improve the recognition accuracy when added into these SOTAs.


\section{Conclusion}
In this paper, we propose a spectral nonlocal (SNL) block for capturing the long-range dependencies between spatial pixels (image classification) or temporal frames (video classification). Inspired by the nonlocal-based blocks, we interpret and extend them in the graph view, and design the SNL block, which is more robust and effective. Experimental results demonstrate the clear-cut improvements across image classification, fine-grained image classification, action recognition, and person re-identification tasks. 
\bibliography{example_paper}
\bibliographystyle{icml2020}

\clearpage
\appendix
\twocolumn[
\icmltitle{Appendix}
]
\section{Graph Fourier transform and graph filter}
\label{appendix_gft}
The classical Fourier transform is defined as:
\begin{equation}
    \hat{F}(\xi) := \left \langle F,  e^{-2 \pi i \xi t} \right \rangle = \int_{R}F(t) e^{-2 \pi i \xi t} dt
\end{equation}

is the expansion of a function $f$ in terms of the complex exponentials. 

Analogously, the graph Fourier transform $\hat{\mathscr{F}}$ of any function $\mathscr{F} \in \mathbb{R}^{N}$ on the vertices of a graph $\mathcal{G}$ is defined as the expansion of $\mathscr{F}$ in terms of the eigenvectors of the graph Laplacian:
\begin{equation}
    \hat{\mathscr{F}}(\lambda_{l}) := \left \langle \mathscr{F},  u_{l} \right \rangle = \sum^N_{i=1} \mathscr{F}(i) u^{*}_l(i)
\end{equation}
where $\mathbf{\lambda}=\{\lambda_1, \lambda_2, ..., \lambda_l, ...\}$ and $\mU=\{\mathbf{u}_1, \mathbf{u}_2, ... \mathbf{u}_l, ...\}$ are the eigenvalue and eigenvector of the graph Laplacian, $\mathbf{u}^{*}_l$ is the $l_{th}$ column vector of $\mU^{\top}$. 

The inverse graph Fourier transform $\hat{\mathscr{F}}^{-1}$ then given by
\begin{equation}
    \hat{\mathscr{F}}^{-1}(\lambda_{l}) = \sum^N_{i=1} \mathscr{F}(i) u_l(i)
\end{equation}

Based on the graph Fourier transform, the graph convolution of the input signal $\mathbf{x}$ with a filter $\mathbf{g}_{\theta}$ can be defined as

\begin{equation}
    \mathbf{x} *_{\mathcal{G}} \mathbf{g}_{\theta} =  \mathscr{F}^{-1}(\mathscr{F}(\mathbf{x}) \odot \mathscr{F(\mathbf{g})})
\end{equation}

where $\odot$ the Hadamard product. If we denote a filter as $\mathbf{g}_{\theta} = diag(\mU^{\top})$

\section{Details of the Proving for the relations}
\label{appendix_prove_nl}

We give the details of the relationship between other nonlocal operators in the spectral view discussed in our paper (Sec.3.2). In the following proving, we assume that $X \in \mathbb{R^{N \times C}}$, $\mZ = g(\mX) = \mX \mathbf{W}_Z$, $\mM_{ij} = f(\mX_i, \mX_j)$. All the normalized term uses the inverse of the degree $1 / d_i$ where $d_i = \sum_j f(\mX_i, \mX_j)$. We also merge the output of the operators with the weight kernel $\mathbf{W} \in \mathbb{R}^{N \times C}$ and defines it as $\mO$ for consistency. Thus the target formulations in this section are a bit different with the definition in their own papers.

\subsection{Nonlocal Block}

The Nonlocal (NL) Block in the spectral view is the same as defining the graph $\mathcal{G}=(\mathbb{V}, \mD^{-1} \mM, \mZ)$ and then using the second term of the Chebyshev Polynomail to approximate the graph filter.

\begin{proof}
The target is to meet the definition of the nonlocal mean operator given in the Eq.(1) in \citet{nl}:
\begin{equation}
\mO_{i,:} =  \frac{\sum_j \Big[f(\mX_{i, :}, \mX_{j, :}) g(\mX_{j, :}) \Big] }{\sum_j f(\mX_{i,:}, \mX_{j,:})} \mathbf{W}
\end{equation}

We firstly define the graph $\mathcal{G}=(\mathbb{V}, \mA, \mZ)$ to represent the graph structure of the nonlocal operator, where the affinity matrix $A$ is calculated by:

\begin{equation}
    \mA = \mD^{-1}_M \mM, \quad \mM = f(\mX_{i,:}, \mX_{j,:})
\end{equation}

Thus, each element of the affinity matrix $\mA$ is:
\begin{equation}
    \mA_{ij}= (\mD^{-1}_M \mM)_{ij} = \frac{f(\mX_{i,:}, \mX_{j,:})}{\sum_j f(\mX_{i,:}, \mX_{j,:})}
    \label{nl_elementA}
\end{equation}

The graph filter $\mathbf{\Omega}$ on $\mathcal{G}$ is approximated by the Chebyshev Polynomail (Eq.~(8) in our paper). When only choosing the second term, it becomes :

\begin{equation}
    F(\mA, \mZ) = \mA \mZ \mathbf{W}
\end{equation}

Then taking Eq.~(\ref{nl_elementA}) into this equation, it becomes:
\begin{equation}
    F_{i,:}(\mA, \mZ) = \frac{\sum_j \Big[f(\mX_{i, :}, \mX_{j, :}) g(\mX_{j, :}) \Big] }{\sum_j f(\mX_{i,:}, \mX_{j,:})} \mathbf{W}
\end{equation}

\end{proof}

\subsection{Nonlocal Stage}

The Nonlocal Stage (NS) in the spectral view is the same as defining the graph $\mathcal{G}=(\mathbb{V}, \mD^{-1}_{M} \mM, \mZ)$ and then using the $1_{st}$-order Chebyshev Polynomail to approximate the graph filter with the condition $\mathbf{W}_1 = \mathbf{W}_2 = -\mathbf{W}$.

\begin{proof}
The target is to meet the definition of the nonlocal stage operator given in the Eq.~(8) in \citet{ns} when merging it with the weight kernel $\mathbf{W} \in \mathbb{R}^{N \times C}$:
\begin{equation}
\mO_{i,:} = \frac{\sum_j \Big[f(\mX_{i,:}, \mX_{j,:}) (\mZ_{j,:} - \mZ_{i,:})\Big]}{\sum_j f(\mX_{i,:}, \mX_{j,:})} \mathbf{W}
\end{equation}

Similar with the proof of NL, we can get each element of the affinity matrix $\mA$ as:
\begin{equation}
    \mA_{ij}= (\mD^{-1}_M \mM)_{ij} = \frac{f(\mX_{i,:}, \mX_{j,:})}{\sum_j f(\mX_{i,:}, \mX_{j,:})}
    \label{ns_elementA}
\end{equation}

The graph filter $\bm{\Omega}$ on $\mathcal{G}$ is approximated by the Chebyshev Polynomail (Eq.~(8) in our paper). When using the $1_{st}$-order Chebyshev Approximation, it becomes:

\begin{equation}
    F(\mA, \mZ) = \mZ \mathbf{W}_1 - \mA \mZ \mathbf{W}_2
\end{equation}

When sharing the weight for $\mathbf{W}_1$ and $\mathbf{W}_2$, i.e $\mathbf{W}_1 = \mathbf{W}_2 = -\mathbf{W}$, we get:

\begin{equation}
    F(\mA, \mZ) = \mA \mZ \mathbf{W} - \mZ \mathbf{W} 
\end{equation}

Then, taking it into element-wise with $\mZ = g(\mX) = \mX \mathbf{W}_Z$ and Eq.\ref{ns_elementA}, this formulation becomes:
\begin{equation}
    F_{i,:}(\mA, \mZ) = \frac{\sum_j \Big[f(\mX_{i,:}, \mX_{j,:}) \mZ_{j,:} \mathbf{W} \Big]}{\sum_j f(\mX_{i,:}, \mX_{j,:})} - \mZ_i \mathbf{W}
\end{equation}

Due to the fact that $\frac{\sum_j f(\mX_{i,:}, \mX_{j,:})}{\sum_j f(\mX_{i,:}, \mX_{j,:})} = 1$, we can get:

\begin{align}
    F_i(\mA, \mZ) &= \frac{\sum_j \Big[f(\mX_{i,:}, \mX_{j,:}) \mZ_{j,:} \mathbf{W} \Big]}{\sum_j f(\mX_{i,:}, \mX_{j,:})} \nonumber\\
    &\quad\quad\quad\quad\quad\quad - \frac{\sum_j f(\mX_{i,:}, \mX_{j,:})}{\sum_{j,:} f(\mX_{i,:}, \mX_{j,:})} \mZ_{i,:} \mathbf{W} \nonumber\\
    & =  \frac{\sum_j \Big[f(\mX_{i,:}, \mX_{j,:}) (\mZ_{j,:} - \mZ_{i,:})\Big]}{\sum_j f(\mX_{i,:}, \mX_{j,:})} \mathbf{W}
\end{align}

\end{proof}

\subsection{Double Attention Block}

The Double Attention Block in the spectral view is the same as defining the graph $\mathcal{G}=(\mathbb{V}, \overline{\bm{M}}, \mZ)$ and then using the second term of the Chebyshev Polynomail to approximate the graph filter, i.e $F(\mA, \mZ) = \overline{\bm{M}} \bm{Z} \mathbf{W}$:

\begin{proof}
The target is to meet the definition of the $\mathrm{A^2}$ operator given in the Eq.(7) in \citet{a2net}:

\begin{equation}
\mO = \sigma(\theta(\mX))\sigma(\phi(\mX)^{T} g(\mX)) = f^a(\mX_i, \mX_j) \mX \mathbf{W}
\end{equation}

The difference between the double $\mathrm{A^2}$ operator and the NL operator is only the kernel function that calculating the affinity matrix \cite{a2net}. Thus we can use the similar proving strategy to reformulate the graph view of $\mathrm{A^2}$ operator into the spectral view as in A.1.

\end{proof}

\subsection{Compact Generalized Nonlocal Block}

When grouping all channels into one group, the Compact Generalized Nonlocal Block in the spectral view is the same as defining the graph $\mathcal{G}=(\mathbb{V}^f, \mD^{-1}_{M^f} \mM^f, \text{vec}(\mZ))$ and then using the second term of the Chebyshev Polynomail to approximate the graph filter, i.e $F(\mA, \mZ) =  \mD^{-1}_{M^f} \mM^f \text{vec}(\mZ) W$. Note that due to the dimension of the input feature $\text{vec}(\mZ) \in \mathbb{R}^{NC \times 1}$ which is different with other nonlocal operators, here we uses $\mM^f , \mA^f \in \mathbb{R}^{NC \times NC}$ for clearity.  

\begin{proof}
The target is to meet the definition of the nonlocal means operator given in the Eq.(7-8, 13) in \citet{cgnl} when merging it with the weight kernel $\mathbf{W} \in \mathcal{R}^{N \times C}$:
\begin{equation}
\text{vec}(\mO) = f(\text{vec}(\mX), \text{vec}(\mX)) \text{vec}(\mZ) \mathbf{W}
\end{equation}

For simplirity, we use $\bm{x}$ to represent $\text{vec}(\mX)$, thus the target becomes:

\begin{equation}
    \bm{o} = f(\bm{x}, \bm{x}) \bm{z} \mW
\end{equation}

Then, we define the graph $\mathcal{G}=(\mathbb{V}^f, \bm{A}^f, \bm{z})$ to define the nonlocal operator in the spectral view, where the set $\mathbb{V}^f$ contains each index (including position and channel) of the vector $\bm{x}$. The affinity matrix $\bm{A}$ is calculated by:

\begin{equation}
    \mA^f = \mM^f, \quad \mM^f = f(\bm{x}, \bm{x})
\end{equation}

The graph filter $\bm{\Omega}$ on $\mathcal{G}$ is approximated by the Chebyshev Polynomail. When only choosing the second term, it becomes:

\begin{equation}
    F(\mA^f, \bm{z}) = \mA^f \bm{z} \mathbf{W} = f(\bm{x}, \bm{x}) \bm{z} \mathbf{W}
\end{equation}

\end{proof}

\subsection{Criss-Cross Attention Block}

The Criss-Cross Attention Block in the spectral view is the same as defining the graph $\mathcal{G}=(\mathbb{V}, \mD^{-1}_{\mC \odot \mM} \mC \odot \mM, \mX)$ and then using the second term of the Chebyshev Polynomail to approximate the graph filter with node feature $\mX$:

\begin{proof}
The target is to meet the definition of the criss-cross attention operator given in the Eq.~(2) in \citet{ccnet}:
\begin{align}
\mO_{i,:} &= \sum_{j \in {\mathbb{V}^{i}}} \mA_{ij} \bm{\Phi}_{j,:} = \frac{\sum_{j \in \mathbb{V}^{i}} f(\mX_{i,:}, \mX_{j,:}) \mX_{j,:} }{\sum_{j \in \mathbb{V}^{i}} f(\mX_{i,:}, \mX_{j,:})} \mathbf{W} \nonumber
\end{align}

in which the set $\mathbb{V}^{i}$ is collection of feature vector in $\mathbb{V}$ which are in the same row or column with position $u$.

Then, we define the graph $\mathcal{G}=(\mathbb{V}, \widetilde{\bm{A}}, \bm{X})$ to represennt the criss-cross attention operator in the spectral view. The affinity matrix $\widetilde{\bm{A}}$ is calculated by:

\begin{equation}
    \begin{split}
    &\widetilde{\bm{A}} = \bm{D}^{-1}_{\mC \odot \mM} \mC \odot \mM , \quad \mM = f(\mX_i, \mX_j) \\
    &   \mC_{ij} =  
        \begin{cases}
        1& j \in \mathbb{V}^{i} \\
        0& \text{else}
        \end{cases} , \nonumber
    \end{split}
\end{equation}

We use $\widetilde{\bm{M}}$ to represent $\mC \odot \mM$, i.e. $\widetilde{\bm{M}} = \mC \odot \mM$. Thus, each element of the affinity matrix $\widetilde{\bm{M}}$ is:
\begin{equation}
    \begin{split}
       \widetilde{\bm{M}}_{ij} =  
        \begin{cases}
        \mM_{ij}& j \in \mathbb{V}^{i} \\
        0& \text{else}
        \end{cases} , \nonumber
    \end{split}
\end{equation}

Thus, we can get the defination of each element in the affinity matrix $\widetilde{\bm{A}}$:

\begin{equation}
    \begin{split}
       \widetilde{\bm{A}}_{ij} =  
        \begin{cases}
        \frac{f(\mX_i, \mX_j)}{\sum_{j \in \mathbb{V}^{i}} f(\mX_i, \mX_j)}, & j \in \mathbb{V}^{i} \\
        0,& \text{else}
        \end{cases} , \nonumber
    \end{split}
\label{cc_elementA}
\end{equation}

The graph filter $\bm{\Omega}$ on $\mathcal{G}$ is approximated by the Chebyshev Polynomail (Eq.~8 in our paper). When using the second term to approximate, it becomes:

\begin{equation}
    F(\widetilde{\bm{A}}, \mX) = \widetilde{\bm{A}} \mX \mathbf{W}
\end{equation}

When taking Eq.\ref{cc_elementA} into this formulation, we can get:
\begin{equation}
    \begin{split}
    	F_{i,:}(\widetilde{\bm{A}}, \mX) &= (\frac{\sum_{j \in \mathbb{V}^{i}} f(\mX_{i,:}, 	\mX_{j,:}) \mX_{j,:}}{\sum_{j \in \mathbb{V}^{i}} f(\mX_{i,:}, \mX_{j,:})}  + \sum_{j \notin \mathbb{V}^{i}} 0 \mX_{j,:}) \mathbf{W}\\
	&=  \frac{\sum_{j \in \mathbb{V}^{i}} f(\mX_{i,:}, 	\mX_{j,:}) \mX_{j,:} }{\sum_{j \in \mathbb{V}^{i}} f(\mX_{i,:}, \mX_{j,:})} \mathbf{W}
    \end{split}
\end{equation}

\end{proof}

\section{External Experiment on Action Recognization}
\label{appendix_stoa}

For Pseudo 3D Convolutional  Network (P3D) and Motion-augmented RGB Stream (MARS), our SNL block are inserted into the Pseudo- 3D right before the last residual layer of the $res3$. For the Video Transformer Network (VTN), we replace its multi-head self-attention blocks (paralleled-connected NL blocks) into our SNL blocks. We use the model pre-trained on Kinetic dataset and fine-tuning on the UCF-101 dataset. Other setting such as the learning rate and training epochs are the same as the experiment on I3D in our paper.

We can see that all the performance are improved when adding our proposed SNL model especially when training end-to-end on the small-scale dataset. In sum, our SNL blocks have shown superior results across three SOTAs (the VTN and MARS) in the action recognition tasks ($0.30\%$ improvement with VTN, $0.50\%$ improvement with MARS). 

\begin{table}[!htbp]
\centering
\small
\caption{Experiments with state-of-the-art backbone}
\begin{tabular}{ccc}
\toprule
Models & Top1(\%) \\
\hline
P3D & $ 81.23	^{\uparrow0.00}$  \\
P3D + \textbf{SNL} & $ 82.65^{\uparrow1.42}$  \\
\hline
VTN & $ 90.06^{\uparrow0.00}$  \\
VTN + \textbf{SNL} & $ 90.34^{\uparrow0.30}$  \\
\hline
MARS & $ 92.29^{\uparrow0.00}$ \\
MARS + \textbf{SNL} & $ 92.79^{\uparrow0.50}$  \\
\bottomrule
\end{tabular}
\end{table}

\section{Experiment on Video Person Re-identification}
\label{appendix_reid}
\begin{table*}[!htbp]
\centering
\small
\caption{Experiments on Video-Person Reidentification}
\label{reid}
\begin{tabular}{lcc|lcc|lcc}
\toprule
\multicolumn{3}{c|}{Mars} & \multicolumn{3}{c|}{ILID-SVID} & \multicolumn{3}{c}{PRID-2011}\\
\hline
Models & Rank1(\%) & mAP(\%) & Models & Rank1(\%) & mAP(\%)  & Models & Rank1(\%) & mAP(\%) \\
\hline
ResNet50tp & $ 82.30^{\uparrow0.00}$ & $ 75.70^{\uparrow0.00}$ & 
ResNet50tp & $ 74.70^{\uparrow0.00}$ & $ 81.60^{\uparrow0.00}$ &
ResNet50tp & $ 86.50^{\uparrow0.00}$ & $ 90.50^{\uparrow0.00}$ \\

+ NL & $ 83.21^{\uparrow0.91}$ & $ 76.54^{\uparrow0.84}$ & 
+ NL & $ 75.30^{\uparrow0.60}$ & $ 83.00^{\uparrow1.40}$ &
+ NL & $ 85.40^{\downarrow1.10}$ & $ 89.70^{\downarrow0.80}$ \\

\hline
+ \textbf{SNL} & $ 83.40^{\uparrow1.10}$ & $ 76.80^{\uparrow1.10}$ & 
+ \textbf{SNL} & $ 76.30^{\uparrow1.60}$ & $ 84.80^{\uparrow3.20}$ &
+ \textbf{SNL} & $ 88.80^{\uparrow2.30}$ & $ 92.40^{\uparrow1.90}$ \\
\bottomrule
\end{tabular}
\end{table*}

For the backbone, we follow the strategy of \citet{reid_tp} that use the pooling (RTMtp) to fuse the spatial-temporal features. (Note that the models are totally trained on ilidsvid and prid2011 rather than fintuning the pretrained model on Mars.) We only insert the SNL block into the ResNet50 (right before the last residual block of \textit{res4}) in RTMtp. Other block setting are the same as the setting for fine-grained image classification on CUB in our paper. For all those datasets we train the model with Adam with the initial learning rate $3e-4$, the weight decay $5e-4$. The learning rate is divided by $200$ at $400$ epochs. All the models are trained for $500$ epochs with the Cross Entropy and  Triplet Loss($margin = 0.3$). We use rank-1 accuracy (Rank1) and men average precision (mAP) to evaluate the performance for the models.

For the large scale dataset Mars\cite{mars_reid} which contains
$1261$ pedestrians captured by at least 2 cameras. The bounding boxes are generated by the detection algorithm DPM and tracking algorithm GMMCP, which forms 20715 person sequences. From Table.~\ref{reid} (Mars), we can see that our SNL can generate $1.10\%$ improvement both on Rank1 and mAP than the backbone, which are both higher than the original nonlocal block ($0.91\% on Rank1$, $0.84\%$ on mAP).

We also generate experiments on two relatively small datasets: ILID-SVID datasets which contains $300$ pedestrians captured by two cameras with $600$ tracklets; PRID-2011 dataset which contains $200$ pedestrians captured by two cameras with $400$ tracklets. From Table.~\ref{reid} (ILID-SVID), we can see that our model can generate $1.60\%$ and $3.20\%$ improvement on the Rank1 and mAP respectively for the ILID-SVID dataset. Moreover, on PRID-2011, we get a more higher improvement ($2.30\%$ on Rank1, $1.90\%$ on mAP) as shown in Table.~\ref{reid} (PRID-2011).

\end{document}